\def\year{2022}\relax
\documentclass[letterpaper]{article} 
\usepackage{aaai22}  
\usepackage{times}  
\usepackage{helvet}  
\usepackage{courier}  
\usepackage[hyphens]{url}  
\usepackage{graphicx} 
\usepackage{natbib}  
\usepackage{caption} 
\usepackage{subcaption}
\usepackage{amsmath}
\usepackage{amssymb}
\usepackage{mathtools}

\DeclareCaptionStyle{ruled}{labelfont=normalfont,labelsep=colon,strut=off} 
\usepackage{algorithm}
\usepackage{algorithmic}
\usepackage{array}
\newcolumntype{P}[1]{>{\centering\arraybackslash}p{#1}}
\newcolumntype{M}[1]{>{\centering\arraybackslash}m{#1}}

\usepackage{newfloat}
\usepackage{listings}
\floatstyle{ruled}
\newfloat{listing}{tb}{lst}{}
\floatname{listing}{Listing}
\title{Self-supervised Enhancement of Latent Discovery in GANs}
\author{
    Silpa Vadakkeeveetil Sreelatha\textsuperscript{\rm 1}\equalcontrib, Adarsh Kappiyath\textsuperscript{\rm 2}\equalcontrib, S Sumitra \textsuperscript{\rm 3}
}
\affiliations{
    \textsuperscript{\rm 1}TCS Research\\
    \textsuperscript{\rm 2}Flytxt Mobile Solutions\\
    \textsuperscript{\rm 3}Indian Institute of Space Science and Technology\\

    silpavs.43@gmail.com, kadarsh22@gmail.com, sumitra@iist.ac.in

}

\usepackage{bibentry}

\begin{document}

\maketitle

\begin{abstract}
Several methods for discovering interpretable directions in the latent space of pre-trained GANs have been proposed. Latent semantics discovered by unsupervised methods are relatively less disentangled than supervised methods since they do not use pre-trained attribute classifiers. We propose Scale Ranking Estimator (SRE), which is trained using self-supervision. SRE enhances the disentanglement in directions obtained by existing unsupervised disentanglement techniques. These directions are updated to preserve the ordering of variation within each direction in latent space. Qualitative and quantitative evaluation of the discovered directions demonstrates that our proposed method significantly improves disentanglement in various datasets. We also show that the learned SRE can be used to perform Attribute-based image retrieval task without further training. 
\end{abstract}

\section{Introduction}
Generative Adversarial Networks (GAN) are generative models that have witnessed significant performance improvements in image synthesis over the last decade \cite{NIPS2014_5ca3e9b1}. It has many applications, including image, audio, and video generation, image manipulation and editing, image-to-image translation, and many others.

The latent space of GANs is hard to interpret due to its high dimensional and abstract structure.  Various architectures such as InfoGAN \cite{infogan}, Structured GAN \cite{NIPS2017_6979}, and many others learn interpretable and meaningful representations from images by either maximizing the information or promoting independence between the latent variables. The fundamental drawback of these approaches is that they fail in the case of complex datasets since the generation quality degrades as they learn to disentangle. To alleviate this problem, recent works such as \cite{shen2020interpreting}, \cite{voynov2020unsupervised} discover interpretable directions directly from the latent space of pre-trained GANs. \cite{voynov2020unsupervised} performs unsupervised learning to identify distinguishable directions while  \cite{NEURIPS2020_6fe43269}, \cite{DBLP:journals/corr/abs-2012-05328} and  \cite{shen2021closedform} obtains directions analytically. These directions need not be completely disentangled.

 We propose Scale Ranking Estimator(SRE), a model learned via \textbf{self-supervision} strategy to enhance disentanglement in the directions derived by current posthoc disentanglement approaches. Self-supervision is a successful training paradigm for deep learning models that allows them to learn in a label-efficient manner. In essence, SRE enhances disentanglement by enforcing the order of variation within each transformation. Our method is independent of the GAN architecture used. We perform extensive qualitative and quantitative analysis on synthetic and natural datasets to show that the proposed method improves the disentanglement of existing directions. SRE learns to encode the magnitude of variation in each direction. We demonstrate a practical application where these encodings can be directly used for Attribute-based image retrieval task.

 \begin{figure*}[t]
\centering
\includegraphics[width=0.95\textwidth]{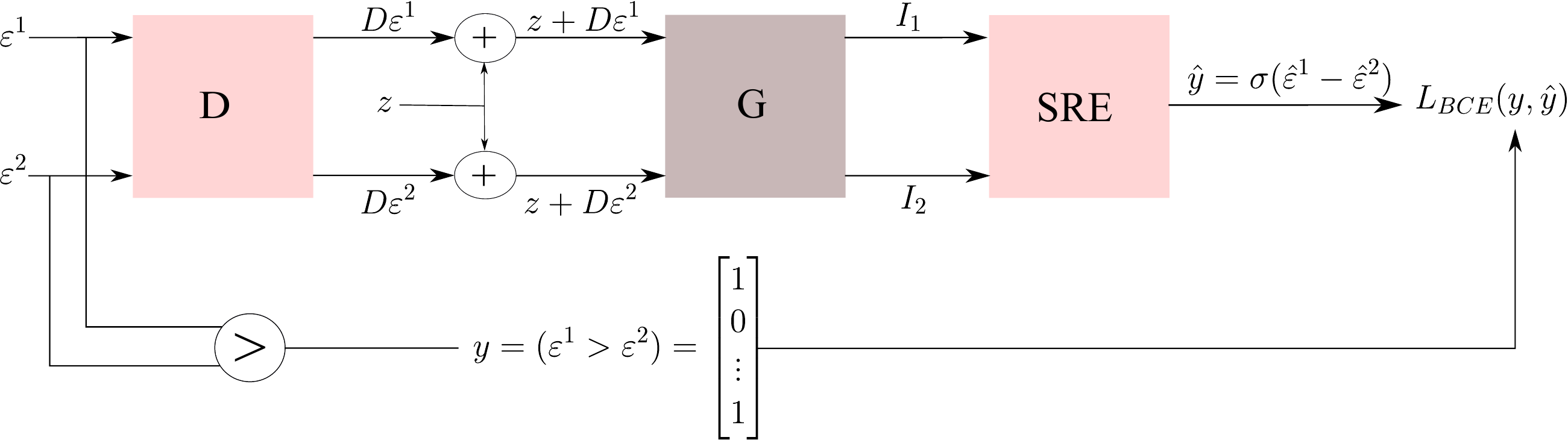} 
\caption{Illustration of the proposed approach; $D$ is initialized with existing post-hoc disentanglement directions. We first compute two linear combinations of directions in $D$, where the coefficients are values in scale vectors $\varepsilon^1$ and $\varepsilon^2$, respectively. These linear combinations are then added to latent code $z$, which gives a pair of shifted latent codes. Generator $G$ outputs a pair of images which are passed to the $SRE$. $SRE$ decodes the scale vectors, $\hat{\varepsilon}^1$ and $\hat{\varepsilon}^2$ from the pair of images. Binary cross-entropy loss is computed based on the difference between the predicted scale vectors and pseudo-ground truth labels. Pseudo-ground truth labels are the original pairwise ordering between the values in $\varepsilon^1$ and $\varepsilon^2$.}
\label{fig:illustration}
\end{figure*}

\section{Related Work}
\textbf{Generative Adversarial Networks} GANs are one of the most popular generative models that shows promising results on image synthesis \cite{NIPS2014_5ca3e9b1}. It consists of a Generator and Discriminator that learns in an adversarial setting. Recent variants of GANs such as StyleGAN \cite{karras2019stylebased}, StyleGAN-2 \cite{Karras_2020_CVPR}, Progressive GAN \cite{DBLP:journals/corr/abs-1710-10196} and BigGAN \cite{brock2018large} are shown to be very successful in generating high-resolution images. Progressive GAN, a successor of conventional GAN, attempts to generate high-resolution images by progressively growing the generator and discriminator. StyleGAN and StyleGAN-2 learn a mapping network that maps the $z$-space to $w$-space that is more disentangled.

\textbf{Post-hoc Disentanglement from pretrained GANs} \cite{DBLP:conf/iclr/HigginsMPBGBML17},  \cite{NEURIPS2018_b9228e09}, \cite{DBLP:conf/icml/0001TFO20} e.t.c. disentangle factors of variations very well in synthetic datasets, but they fail to do so in complex natural datasets. These classical disentanglement learning techniques improve disentanglement at the cost of generation quality. To overcome this limitation, extensive research has been conducted in the field of learning interpretable directions from pre-trained models. They can be categorized into three based on the learning paradigm used :
\begin{itemize}
\item \textbf{Supervised} : \cite{DBLP:conf/iclr/BauZSZTFT19} computes the agreement between the output of a pre-trained semantic segmentation network and the spatial location of the unit activation map to identify the concept encoded in each unit. 
\cite{shen2020interpreting} and \cite{DBLP:journals/ijcv/YangSZ21} use off-the-shelf classifiers to discover interpretable directions in the latent space. A conditional normalizing flow version of \cite{shen2020interpreting} and \cite{DBLP:journals/ijcv/YangSZ21} is explored in \cite{10.1145/3447648}. The main limitation of the above approaches is that they require pre-trained networks, which may not be available for complex transformations. 

\item \textbf{Unsupervised} : \cite{voynov2020unsupervised} discovers interpretable directions in an unsupervised manner by jointly updating a candidate direction matrix and reconstructor that predicts the perturbed direction. \cite{peebles2020hessian} proposes a regularization term that forces the Hessian of a generative model with respect to its input to be diagonal. However, such methods require training.
\cite{NEURIPS2020_6fe43269} observed that applying PCA on the latent space of Style-GAN and BigGAN retrieves human-interpretable directions. \cite{shen2021closedform} and \cite{DBLP:journals/corr/abs-2012-05328} obtained a closed-form solution by extracting the interpretable directions from the weight matrices of pretrained generators. These methods are computationally inexpensive since they do not require any form of training. \cite{voynov2020unsupervised} and \cite{peebles2020hessian} attempts to learn directions that are easily distinguishable while \cite{shen2021closedform}, \cite{DBLP:journals/corr/abs-2012-05328} and \cite{NEURIPS2020_6fe43269} finds directions of maximum variance. However, none of these approaches ensure that only a single factor of variation gets captured in a transformation. Our method addresses this problem by defining a self-supervision task that promotes disentanglement on directions captured by these methods.

\item \textbf{Self-supervised} : \cite{gansteerability} and \cite{Plumerault2020Controlling} make use of user-specified simple transformations as a source of self-supervision to learn corresponding directions. The main drawback of these approaches is that, user-specified edits are hard to obtain for complex transformations.  Unlike these methods, our method relies on transformations discovered by unsupervised methods and hence can discover a wide variety of disentangled transformations.
\end{itemize}

\section{Proposed Method}

Firstly, we provide the intuition behind our approach. In an entangled transformation, formulating a task that favors the dominant factor of variation will enhance the dominant factor in it. To achieve this, we propose Scale Ranking Estimator (SRE), a neural network that learns to rank the scale of each transformation in generated images. Imposing a ranking on the magnitude of variation in each direction would hopefully force the SRE to distinguish between the factors of variation in the associated transformation and thus capture the dominant factor of variation. The directions could then be updated based on this knowledge. An illustration of the proposed approach is given in Figure \ref{fig:illustration}.
 
We formally define all the components involved in our training scheme. Let $G: Z \rightarrow I$ be the pre-trained generator, where $Z$ is the latent space and $I$ represents the pixel space. Interpretable directions are discovered from the latent space of generator $G$. Let $D\in\mathbb{R}^{k \times d}$ denote the matrix whose columns correspond to interpretable directions in latent space. $k$ and $d$ are the latent space dimensionality and the number of interpretable directions, respectively. We also define a neural network $SRE(i ; \theta), i \in I$ that outputs the scale of transformation corresponding to each direction in $D$. $D$ and $SRE$ are the trainable components in our approach, while the parameters of $G$ are non-trainable.

\subsection{Training scheme}
We initialize $D$ with a set of directions obtained from any post-hoc disentanglement method. A linear walk in the latent space is given by $\hat{z} \rightarrow z + D\varepsilon$, where $D\varepsilon$ is the linear combination of directions in $D$. $\varepsilon = (\varepsilon_1, \varepsilon_2, .., \varepsilon_d) \in \mathbb{R}^{d}$, where  $\varepsilon_{i} \sim \textit{U}(-e,e)$ represents the scale of corresponding direction. We sample $\varepsilon^1$,  $\varepsilon^2 \in \mathbb{R}^{d}$ to  generate images $G(\hat{z_{1}})$ and $G(\hat{z_{2}})$, where $\hat{z_{1}} \rightarrow z + D\varepsilon^{1}$ and $\hat{z_2} \rightarrow z + D\varepsilon^{2}$. These images are then passed to $SRE$ which predicts $\hat{\varepsilon}^{1}$ and $\hat{\varepsilon}^{2}$ based on the information encoded in the generated images. 

The loss function to be minimized is as follows :
\begin{equation}
  \label{eq:loss}
   L = \mathop{\mathbb{E}}_{\substack{z\sim N(0,1)\\
                               \varepsilon^{1}, \varepsilon^{2} \sim \textit{U}(-e,e)}}
        \sum^{d}_{i=1} L_{BCE} (y_{i}, \hat{y}_{i}),
\end{equation}

where,

\begin{equation*}
    \hat{y}_{i} = \sigma(\hat{\varepsilon_{i}}^{1} - \hat{\varepsilon_{i}}^{2}),
\end{equation*}

\begin{equation}
        \label{eq:ground_truth}
         y_{i} =
         \begin{cases}
              1, & \text{if } \varepsilon_{i}^{1} > \varepsilon_{i}^{2},\\
            0,  & \text{otherwise.}
         \end{cases}
\end{equation}

Here, $L_{BCE}$ is the binary cross-entropy loss between the predicted output and the pseudo ground-truth, $y_i$ is determined by comparing the scale of transformation used to generate the images as shown in \eqref{eq:ground_truth}.  We provide self-supervision using the knowledge already present in the initialized direction matrix to update it further. 

We perform weight updates on $D$ and $SRE$ in an alternate fashion. There are two optimization steps in each training iteration. Firstly, we compute the loss as specified in \eqref{eq:loss} to update the weights of $SRE$ by freezing the weights of direction matrix $D$. In the subsequent step, we use the updated $SRE$ to recalculate the loss as in \eqref{eq:loss}. The parameters of $SRE$ are now freezed to update $D$.  Training in this manner helps continually transfer some of the information from $SRE$ to $D$ and vice-versa. This is critical since the initialization of $SRE$ is random, whereas the initialization of $D$ is partially learned directions. 

As discussed above, $\varepsilon$ is sampled from a multivariate uniform distribution with parameters $e$ and $-e$. If the specified range $(-e,e)$ is relatively small, our method becomes highly constrained, making it hard to capture the variations in a disentangled factor. If the range is set very wide, the model has the freedom to allow a lot of variation in the transformation, which can cause it to stay entangled. As a result, determining the correct values for the hyper-parameter $e$ is crucial for improved disentanglement. 

\section{Experimental Details}
This section discusses the datasets used, pre-trained generators corresponding to each dataset, choice of initialization for $D$, and hyperparameters involved.

\begin{figure}
\begin{center}
\includegraphics[width=1.0\columnwidth]{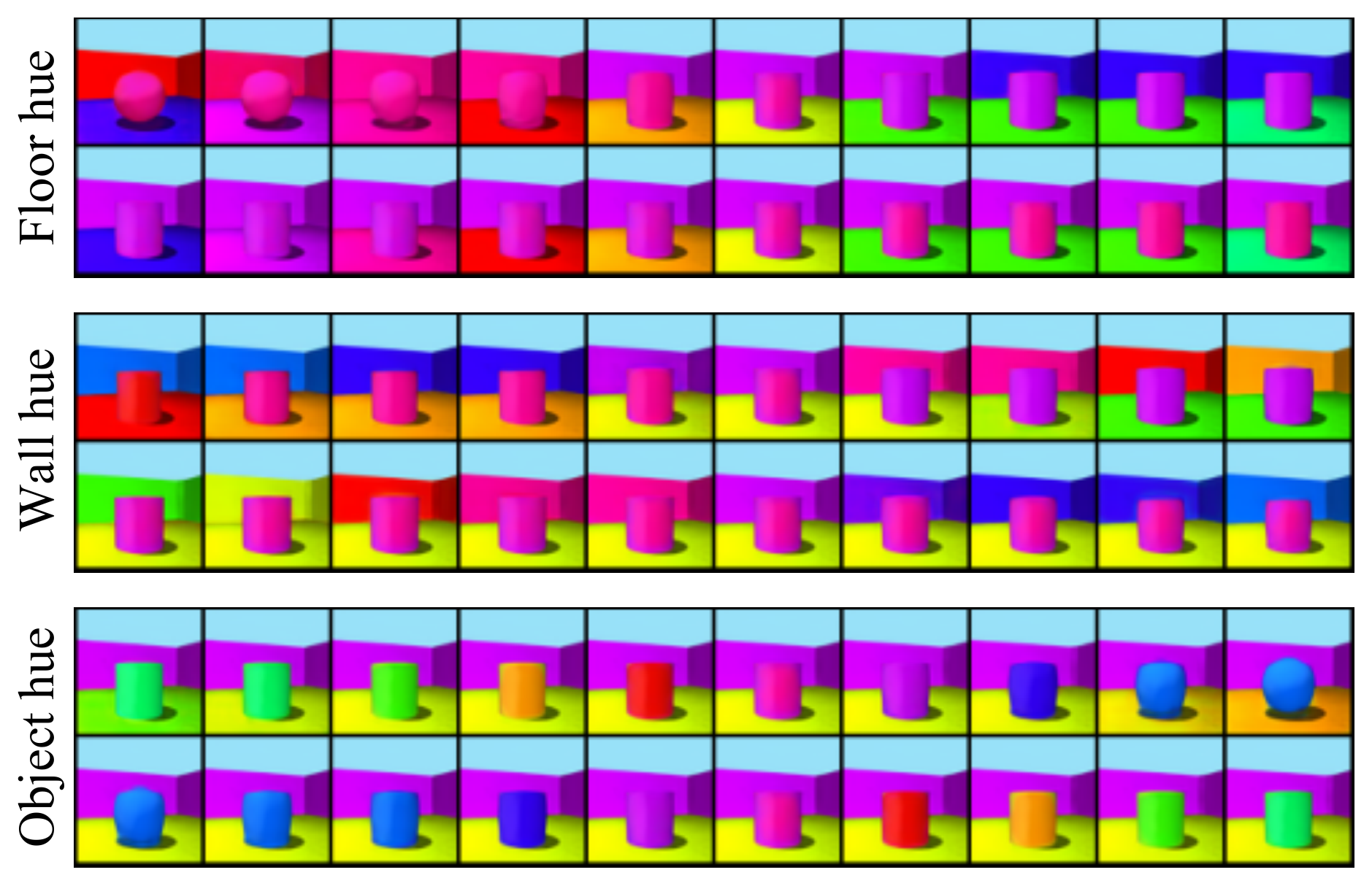}
\end{center}
  \caption{Latent traversal corresponding to Floor hue, Wall hue, Object hue on 3DShapes. For each attribute, the first row corresponds to SeFa and the second row corresponds to SeFa + SRE. }
\label{fig:3dshapes}
\end{figure}

\subsection{Datasets}
We perform the experiments on following datasets:
\begin{itemize}
    \item \textbf{CelebA-HQ} \cite{DBLP:journals/corr/abs-1710-10196} consist of 30,000, $1024 \times 1024$ resolution images of Celebrity faces.
    \item \textbf{AnimeFaces dataset} \cite{DBLP:journals/corr/abs-1708-05509} consist of $64 \times 64 $ resolution face images of Anime characters.
    \item \textbf{LSUN-Cars} \cite{journals/corr/YuZSSX15} consist of $512\times512$ resolution images of cars.
    \item \textbf{3D Shapes} \cite{3dshapes18} containing 480,000 images of $64\times64$ resolution synthetic images with 6 factors of variation. 
\end{itemize}

\subsection{Pre-trained Generators}
We use four different variants of GAN for our experiments to show that our method is independent of the GAN architecture used. \textbf{PGGAN} \cite{DBLP:journals/corr/abs-1710-10196} is used for generating samples from CelebA-HQ dataset. As a representative of conventional GANs, we use \textbf{Spectral Norm GAN} \cite{DBLP:conf/iclr/MiyatoKKY18} to generate Anime Faces. \textbf{StyleGAN}, \cite{karras2019stylebased} and \textbf{StyleGAN-2} \cite{Karras_2020_CVPR} are used for LSUN-Cars and 3DShapes dataset, respectively. We used the same pre-trained generators that are used in \cite{voynov2020unsupervised} and \cite{shen2021closedform}. 

\subsection{Initialization}
We mainly use two contrasting post-hoc disentanglement algorithms to obtain initialization for the direction matrix $D$. As a sampling-based initialisation, we consider \textbf{SeFa} \cite{shen2021closedform}  because it does not require any form of training, whereas the other initialization used is based on directions learned by \textbf{LD} \cite{voynov2020unsupervised}, which requires learning to obtain interpretable directions. We show that our method enhances the disentanglement of any set of directions regardless of the paradigm used to generate it, be it sampling or learning. We used the implementation released by the authors of \cite{voynov2020unsupervised} and \cite{shen2021closedform} to derive the directions for initialization on all the datasets.

 \begin{figure*}[h]
\centering
\includegraphics[width=1.0\textwidth]{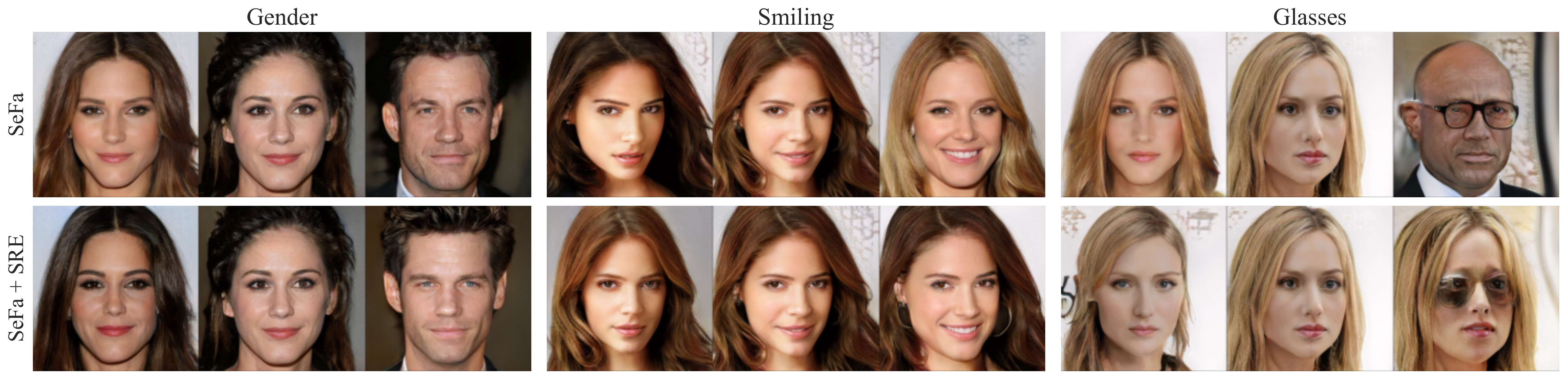} 
\caption{Comparison of latent traversals obtained by SeFa and SeFa + SRE on CelebA-HQ dataset. For each attribute, middle image represents the original image, image to the left and right represents the source image manipulated in positive and negative directions, respectively.}
\label{fig:celebacf}
\end{figure*}

\begin{figure*}[h]
\begin{center}
\includegraphics[width=.3\textwidth]{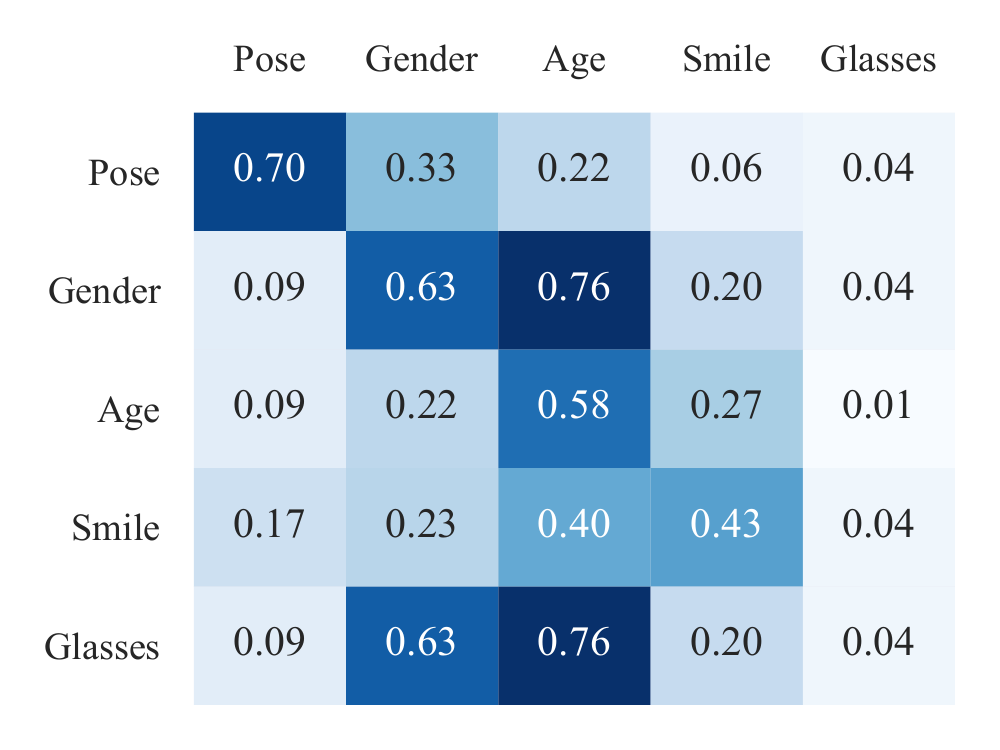}
\includegraphics[width=.3\textwidth]{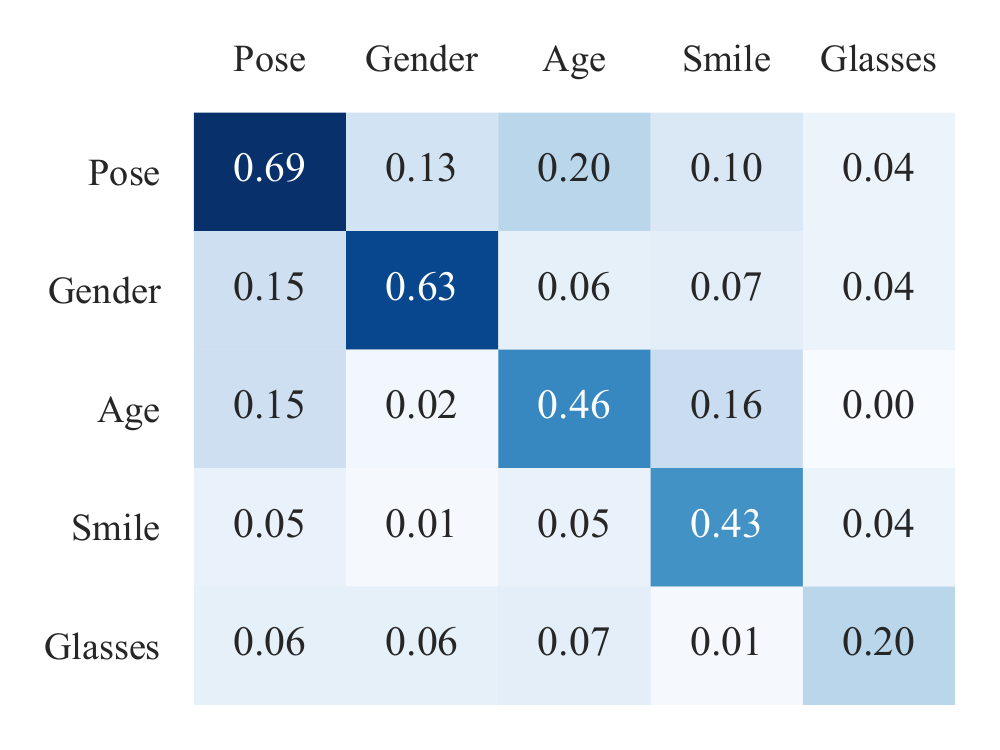}
  \caption{ Rescoring matrix obtained for SeFa (Left) and SeFa + SRE (Right) on CelebA-HQ dataset. Each row represents an attribute obtained by moving in the relevant direction, and the column corresponds to attribute predictors used to compute the scores.}
\label{fig:rescoring}
\end{center}
\end{figure*}

\subsection{Hyperparameters}
\begin{itemize}
    \item \textbf{Architecture} : For all the four datasets, We utilize ResNet-18 model \cite{7780459} for $SRE$ while $D$ is a simple linear operator. We discovered that ensuring orthogonality between directions in $D$ in each iteration resulted in better disentanglement. 
    \item \textbf{Number of iterations} : We set number of iterations to be 6000 for 3DShapes and 20000 for all other datasets. 3DShapes requires relatively lesser number of iterations since it is a synthetic dataset.
    \item  \textbf{Optimization} : We use Adam optimizer to optimize both $D$ and $SRE$. The learning rate is set to 0.0001. Batch size is 64 for 3DShapes and 8 in the case of CelebA-HQ dataset. For all other datasets, batch size is set to 16. 
\end{itemize}

\begin{figure*}[t]
\centering
\includegraphics[width=1.0\textwidth]{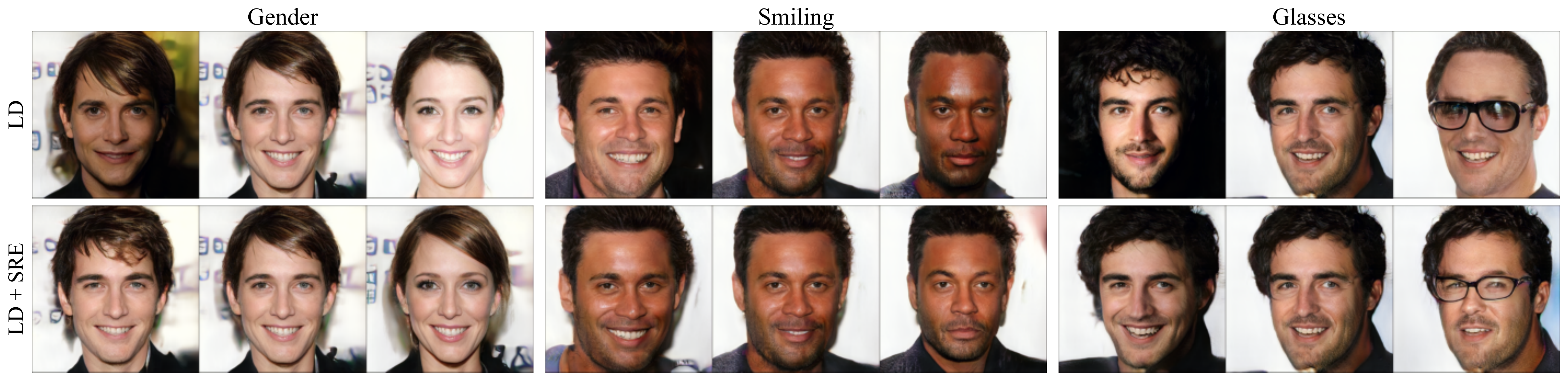} 
\caption{Comparison of latent traversals obtained by LD and LD + SRE on CelebA-HQ dataset. For each attribute, middle image represents the original image, image to the left and right represents the source image manipulated in positive and negative directions, respectively.}
\label{fig:celebald}
\end{figure*}

\begin{figure*}[t]
\centering
\includegraphics[width=1.0\textwidth]{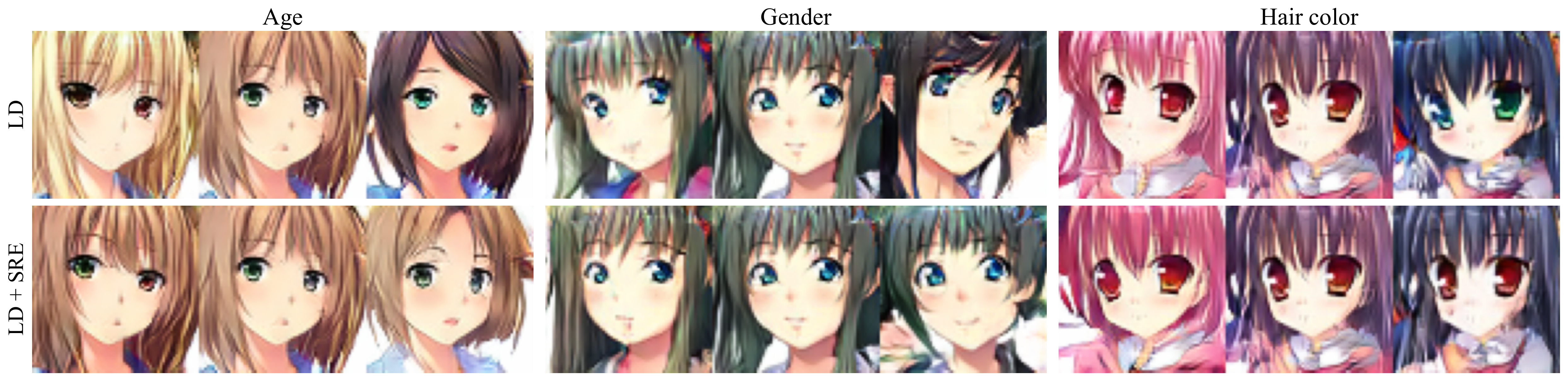} 
\caption{Comparison of latent traversals obtained by LD and LD  + SRE on AnimeFaces dataset. For each attribute, middle image represents the original image, image to the left and right represents the source image manipulated in positive and negative directions, respectively.}
\label{fig:animeld}
\end{figure*}

\section{Results}

In this section, we discuss the qualitative and quantitative results for each of the datasets. SeFa + SRE and LD + SRE correspond to our approach where D is initialized with SeFa and LD directions, respectively. We compare the performance of  SeFa + SRE with SeFa and LD + SRE with LD directions.

\subsection{Qualitative Analysis}
\begin{figure}[t]
\begin{center}
\includegraphics[width=1.0\columnwidth]{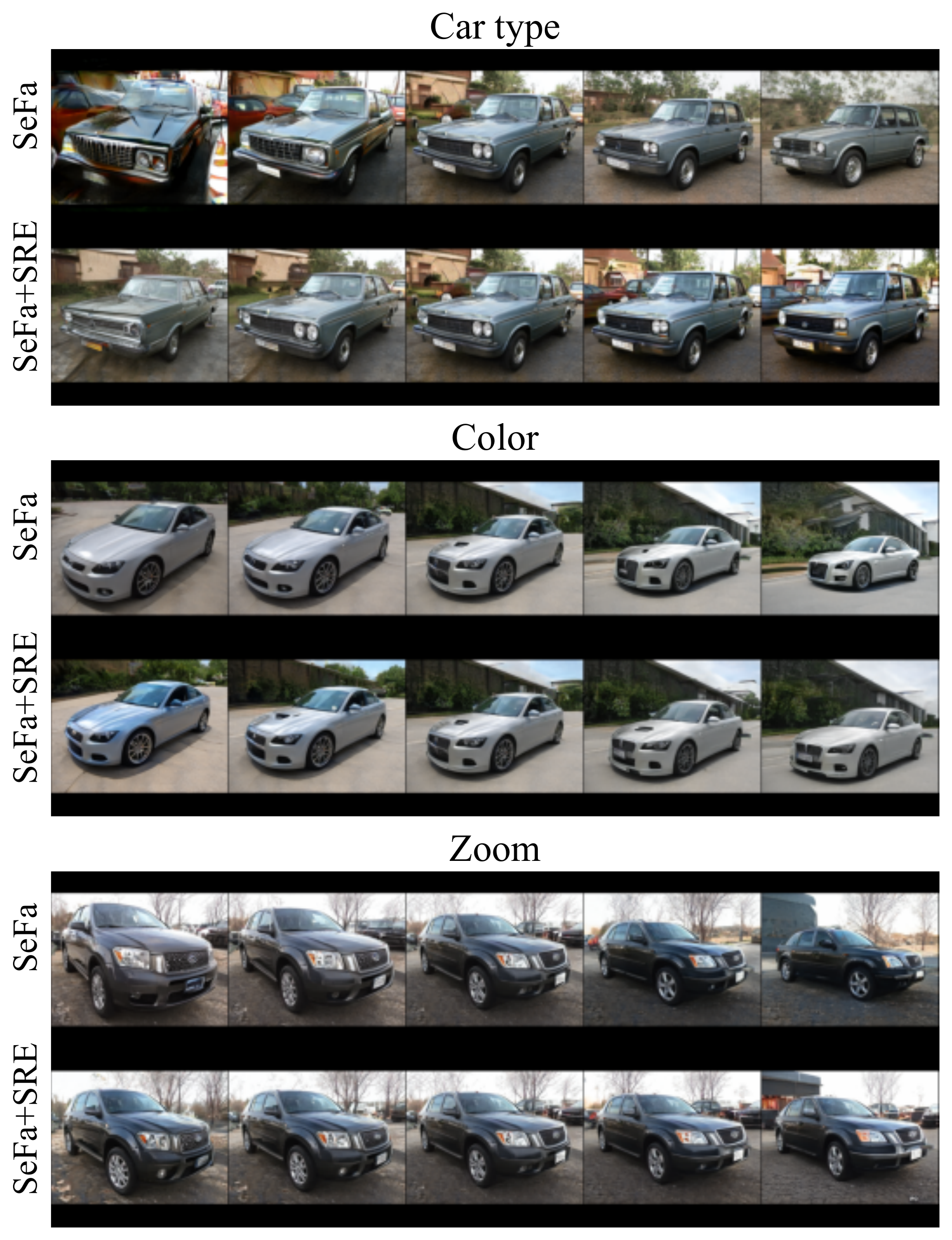}
\end{center}
  \caption{Comparison of latent traversals obtained by SeFa and SeFa + SRE on LSUN Cars dataset. For each attribute, middle image represents the original image.}
\label{fig:carscf}
\end{figure}

We conducted a thorough qualitative analysis to evaluate the performance of our proposed approach. Firstly, we analyze the performance of SeFa compared to SeFa + SRE on 3DShapes, CelebA, and LSUN Cars datasets. We plot the latent traversal starting with the original image by traversing in opposite sides along the relevant directions. The range of $\varepsilon$ is set from -10 to 10. Figure \ref{fig:3dshapes} shows the qualitative results on Shapes3D for three different attributes, floor hue, wall hue, and object hue.  It can be observed that floor hue is entangled with object shape and wall hue in the case of SeFa directions while our method improves these directions by disentangling floor hue from the other attributes. A similar trend can be seen in the case of wall hue which is entangled with floor hue in SeFa. More latent traversals on 3DShapes are available in Technical Appendix. Qualitative analysis on CelebA-HQ and LSUN Cars dataset is demonstrated in Figure \ref{fig:celebacf} and Figure \ref{fig:carscf}. Each transformation corresponding to SeFa is entangled with one or more attributes. In Figure \ref{fig:carscf}, car type and color are entangled with zoom in case of SeFa. However, SeFa + SRE disentangles these attributes from zoom. Similarly, SeFa entangles zoom with orientation whereas our method preserves zoom by removing the effect of orientation. Qualitative analysis shows that SRE applied on SeFa initialization disentangles SeFa directions.

We also analyze the directions obtained by LD and SRE applied on LD initialization as shown in Figure \ref{fig:celebald} and \ref{fig:animeld}. Even though LD seeks distinguishable directions, it can be seen that the transformations obtained are quite entangled on both CelebA-HQ and AnimeFaces datasets. We noticed that these transformations are less identity-preserving which is reflected in the Identity preservation accuracies shown in Table \ref{tab1:celebahq}. Qualitative analysis shows that our approach based on SRE updates the directions so that it results in disentangled and identity-preserving transformations. Additional latent traversal  on various datasets and directions are provided in the Technical Appendix.

\subsection{Quantitative Analysis}
We perform quantitative analysis on CelebA-HQ and 3DShapes to evaluate the proposed approach.
The two quantitative metrics that we employed to analyze the performance on CelebA-HQ dataset are Rescoring Analysis and Identity Preservation accuracy. We use pre-trained attribute predictors released by the authors of \cite{han2021IALS} to perform rescoring analysis. These attribute predictors are binary classifiers trained on each of the 40 attributes of CelebA dataset \cite{liu2015faceattributes}. We perform rescoring analysis similar to that described in \cite{shen2020interpreting}. We take a random sample of 2000 generated images and manipulate them in the direction of the desired attribute.
The pre-trained attribute predictors are then used to obtain predictions for the original and altered images. We subsequently compute the absolute value of the difference between the predictions produced for the original image and the manipulated image. The rescoring for the selected direction is computed by taking the mean of this metric across images. Figure \ref{fig:rescoring}
shows the rescoring matrix corresponding to SeFa and SeFa + SRE. It can be seen that, when applied on SeFa initialization, SRE better disentangles each of the five attributes compared to SeFa. This analysis supports the qualitative analysis discussed in the previous section. The directions updated by SRE retain the knowledge of individual attributes while reducing the entanglement with other attributes. As observed in the rescoring matrix, SeFa fails to capture Eyeglasses. However, there are directions in SeFa that encodes eyeglasses as the dominant factor. By dominant factor, we mean that it is dominant compared to other factors while the magnitude of the variation is less. SRE disentangles eyeglasses better in one of these directions, which is an interesting observation. This shows that SRE can disentangle factors of finer variation that the initialization struggles to capture. We provide a summary of rescoring to compare the performance of our approach with SeFa and LD. Since the rescoring matrix should be close to the diagonal matrix in case of ideal disentanglement, we compute the ratio of the sum of squares of diagonal elements to that of the off-diagonal elements. Higher the value, better the disentanglement. These values are reported in Table \ref{tab1:celebahq}. 

Identity preservation accuracy is also computed to see how effectively SRE retains identity while enhancing disentanglement. We randomly sample 2000 generated images and edit them in the desired direction to obtain the manipulated images. The pair of images are fed into the face recognition model given by \cite{face-recog}, which returns a binary value indicating whether the faces are similar or not. We repeat this procedure in three different directions for all the methods to compute average Identity preservation accuracy. Table \ref{tab1:celebahq} summarizes these values. Results suggest that SRE implicitly learns to preserve identity as it learns to disentangle. We believe that our model learns to incorporate smoothness while learning a ranking function on the scale of transformation which helps it to preserve identity. 

We also perform quantitative evaluation on 3DShapes since the ground truth factors are readily available. We train SRE and the baselines using seven pretrained StyleGAN generators for each random seed. Training is done for five different random seeds. We calculate Mutual Information Gap (MIG) \cite{chen2018isolating} and Factor-VAE \cite{pmlr-v80-kim18b}, which are two widely used disentanglement metrics in the literature. This is done by computing the latent space embeddings for real samples by training a GAN inversion network as in \cite{DBLP:journals/corr/abs-2102-06204}. The evaluation metrics are computed on real samples to analyze the performance of disentangled directions. The mean and standard deviation of the metrics across the models trained for different seeds are reported in Table \ref{tab2:3dshapes}. Both the MIG and Factor-VAE metrics show that SRE outperforms the baselines. Applying SRE on top of SeFa and LD directions increases average MIG by 104.54\% and 50\% respectively. Results for additional metrics such as DCI  \cite{eastwood2018a} and $\beta$-VAE \cite{DBLP:conf/iclr/HigginsMPBGBML17} can be found in the Technical Appendix.

The use of directions derived by post-hoc disentanglement algorithms as initialization is motivated by the fact that some of these (SeFa, GANSpace e.t.c.) are computationally cheap to obtain. However, we also performed extensive experimentation on 3DShapes and CelebA-HQ to evaluate the performance of our approach using random initialization. In comparison to SRE with existing post-hoc disentanglement initialization, we observe that SRE with random initialization requires more iterations to converge and a higher learning rate to optimize the direction matrix D.  As shown in Table \ref{tab3:random}, SRE with random initialization performs reasonably well at the expense of longer training time. We also observe a similar trend in the case of CelebA-HQ (Rescoring value : \textbf{6.722}).

\begin{figure}[h]
\begin{center}
\includegraphics[width=1.0\columnwidth]{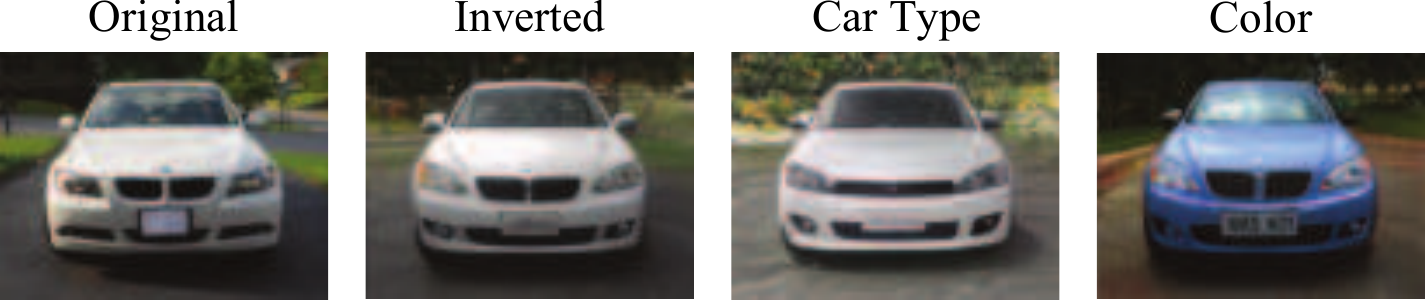}
\end{center}
\caption{Results for Real image editing with respect to multiple attributes in LSUN Cars dataset.}
\label{fig:realedit}
\end{figure}

We perform real image editing using the directions obtained by SRE. We first approximate the latent vector in the latent space for real images using the GAN inversion paradigm mentioned in \cite{10.1145/3450626.3459838} and then shift the latent vector in the direction of desired attributes to obtain the manipulated images. Qualitative results are provided in Figure \ref{fig:realedit} which suggests that the SRE directions can also be applied to real images. 

\begin{table}[h]
\centering
\begin{tabular}{ |M{2.3cm}|M{2.3cm}|M{2.3cm}| }
\hline
Method & Rescoring$(\uparrow)$& Identity Preservation Accuracy$(\uparrow)$\\
\hline
SeFa  & 0.64    &0.61\\
SeFa + SRE&   \textbf{7.77}  & \textbf{0.97}\\
\hline
LD &1.43 & 0.73\\
LD + SRE    & \textbf{3.73} & \textbf{0.94}\\
 \hline
\end{tabular}
\caption{Comparison of Quantitative metrics on CelebA-HQ dataset.}\label{tab1:celebahq}
\end{table}

\begin{table}[h]
\centering
\begin{tabular}{ |M{2.3cm}|M{2.3cm}|M{2.3cm}| }
 \hline
Method & MIG$(\uparrow)$ & Factor-VAE Score$(\uparrow)$\\
 \hline
     SeFa  & 0.22$\pm$0.01    &0.86 $\pm$0.01\\
 SeFa + SRE&   \textbf{0.45$\pm$0.06}  & \textbf{0.94$\pm$0.02}\\
  \hline
    LD &0.14$\pm$0.05 & 0.78$\pm$0.06\\
 LD + SRE    & \textbf{0.21$\pm$0.05} & \textbf{0.90$\pm$0.05}\\
 \hline
\end{tabular}
\caption{Comparison of Quantitative metrics on 3DShapes dataset.}\label{tab2:3dshapes}
\end{table}

\subsection{Effect of epsilon ($\varepsilon$)}
We devise an ablation to study the effect of $\varepsilon$ on training of SRE. We consider three ranges of $\varepsilon $ : (-1, 1), (-3, 3), (-10, 10) with all the other parameters fixed. The results are summarized in Table \ref{tab4:ablation_epsilon}. SRE is able to disentangle factors of variation  with $\varepsilon$ range set to (-1,1). As $\varepsilon $ is progressively increased,  MIG shows a declining trend. Restricting the range of $\varepsilon$ forces the model to accommodate factors that take relatively lesser number of variations, hence forcing the representation to be disentangled. An $\varepsilon$ with larger range provides flexibility to accommodate entangled factors, whereas extremely less range of $\varepsilon$ will not  have sufficient values to properly accommodate variation within a single feature, thus failing to learn any factors of variation properly in both the cases. 

\begin{table}[t]
\centering
\begin{tabular}{ |M{2cm}|M{2cm}|M{1.5cm}| }
 \hline
Initialization & MIG$(\uparrow)$ & Iterations \\
  \hline
 Random  & 0.34 $\pm$ 0.04  & 28000\\
 SeFa &   \textbf{0.45$\pm$0.06} & 6000 \\
 \hline
\end{tabular}
\caption{Quantitative metrics of SRE on different initializations averaged across 5 different random seeds.}
\label{tab3:random}
\end{table}

\begin{table}[t]
\centering
\begin{tabular}{ |M{3cm}|M{3cm}| }
 \hline
Range of $\varepsilon$ & MIG$(\uparrow)$ \\
  \hline
 (-1, 1)  & 0.31\\
 (-3, 3)  & 0.26 \\
 (-10, 10) & 0.18 \\
 \hline
\end{tabular}
\caption{Effect of $\varepsilon$ on performance of SRE.}\label{tab4:ablation_epsilon}
\end{table}

\section{Attribute-based Image Retrieval}
This section demonstrates an immediate practical application of the learned SRE where it can be directly used for Attribute-based image retrieval task. As we already discussed, our approach updates the initialization and enhances disentanglement in the directions. During the training phase, the Scale Ranking Estimator is updated as well to aid the whole learning process. This section explores the possibility of using the trained SRE for  Attribute-based image retrieval without any kind of task specific retraining or fine-tuning.

\begin{figure}[h]
\begin{center}
\includegraphics[width=0.9\columnwidth]{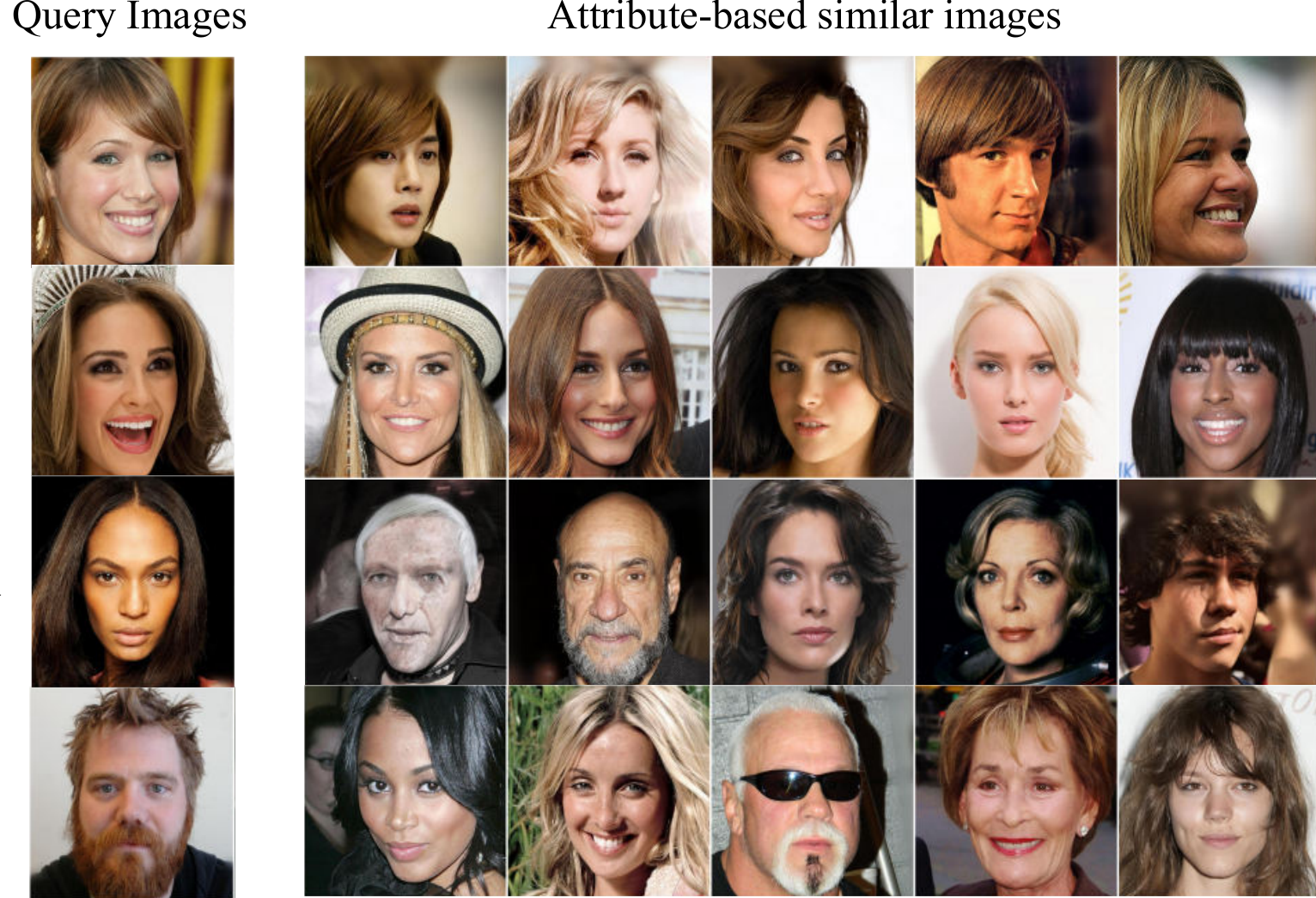}
\end{center}
\caption{Results for Attribute-based image retrieval task on CelebA-HQ dataset. Each row corresponds to attributes Pose, Gender, Expression and Age respectively.}
\label{fig:application}
\end{figure}

Given a query image and a specific attribute, our goal is to retrieve images from a pool of real images similar to the query image based on the specified attribute. Attribute-based image retrieval could be of great use in Reverse image search, Person Identification e.t.c. We provide qualitative evidence to show that the SRE that comes as a by-product of our training process can be used for attribute-based image retrieval without any additional training. During the inference, given any image, SRE  outputs a vector representation where each value at index $i$ represents the scale or the amount of variation of attribute encoded by the direction at index $i$ in the learned direction matrix $D$.   

We first get the output representations from SRE for all the pool images and the query image. For a given attribute, we obtain the attribute-specific variation for all images by accessing the index corresponding to the direction that encodes the given attribute in the output representations. Pool images are sorted in an ascending order based on the Euclidean distance between their attribute-specific variation to that of the query image. Top $K$ images from the sorted set are $K$ images from the pool most similar to the query image with respect to the specified attribute.

We empirically demonstrate the performance of Attribute-based image retrieval on CelebA-HQ data set in Figure \ref{fig:application} . We considered SRE model trained using SeFa initialization for the analysis. We set $K = 5$ for all the attributes. The attributes considered are Pose, Gender, Expression, and Age. According to qualitative results, SRE performs well on the Attribute-based retrieval task, although it is not explicitly trained to do so.

\section{Conclusion \& Future work}
We propose a new method for improving disentanglement and interpretability in the directions obtained by existing post-hoc disentanglement methods by learning the Scale Ranking Estimator (SRE). We also provide a thorough quantitative and qualitative analysis of its performance on various real-world and synthetic datasets. Our approach could be used to improve the disentanglement of any set of existing directions regardless of the underlying algorithm used to obtain them. In addition to enhancing disentanglement, trained SRE can also be used for Attribute-based image retrieval without any task-specific training. 
 Computing a closed-form analytical solution to enforce order on the variation in each transformation would also be helpful to enhance the disentanglement by cutting down the training time. Better quantitative metrics need to be proposed to evaluate post-hoc disentanglement methods on natural datasets which consist of complex attributes.

\bibliography{reference}

\appendix
\onecolumn
\section{Appendix}

We provide latent traversals for 3DShapes, CelebA-HQ and Anime Faces dataset. Each rectangular grid corresponds to one sample. The first row corresponds to latent traversal corresponding to baseline initialization and the second row corresponds to latent traversal in directions obtained after training with SRE. 

\begin{figure*}[h]
\centering
\includegraphics[width=0.7\textwidth]{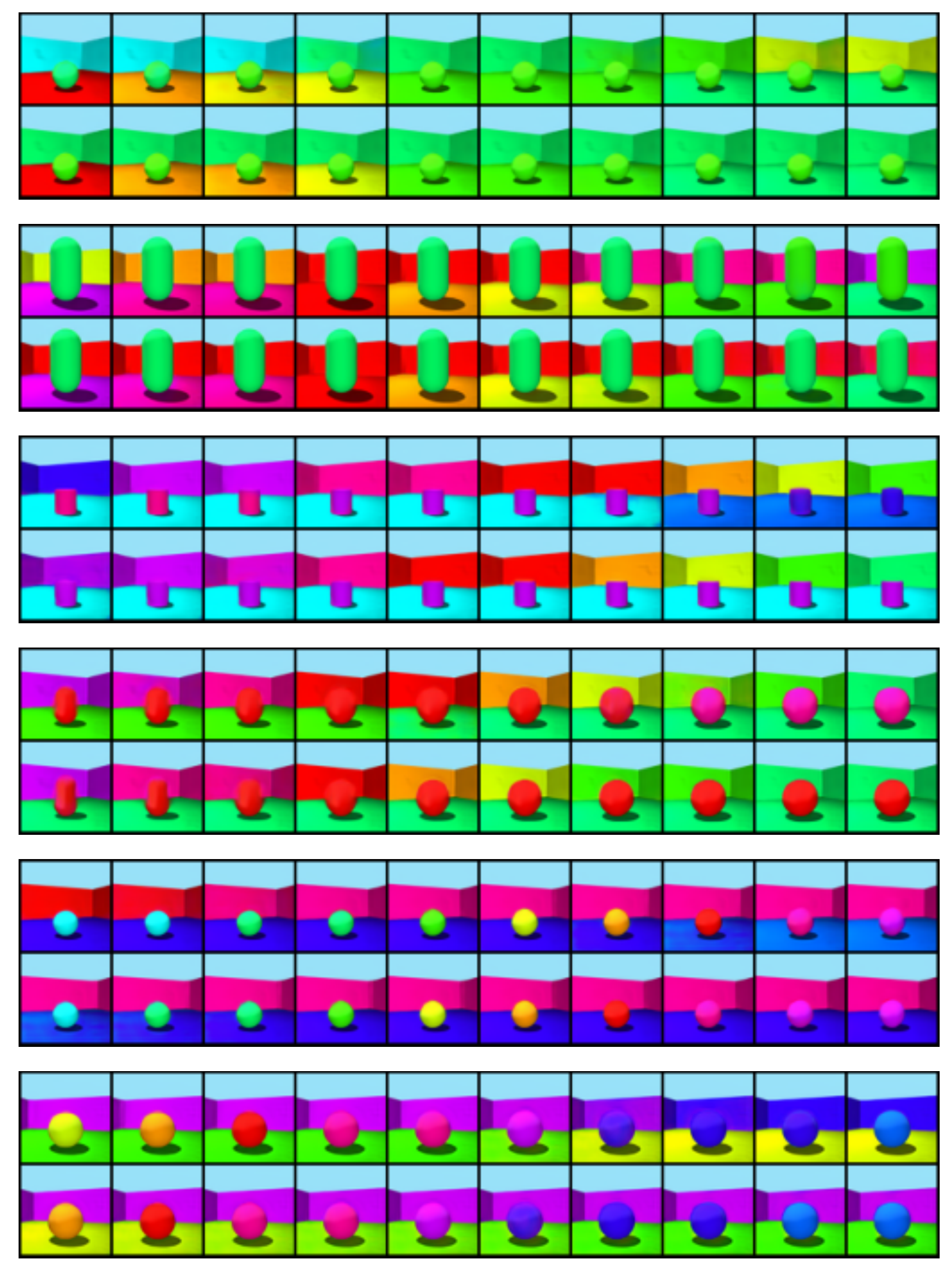} 
\caption{Comparison of latent traversals obtained by SeFa and SeFa + SRE on 3DShapes dataset. Factors shown are Floor hue, Wall hue and Object hue.}
\label{fig:shapes3d_first_three}
\end{figure*}

\begin{figure*}[t]
\centering
\includegraphics[width=0.85\textwidth]{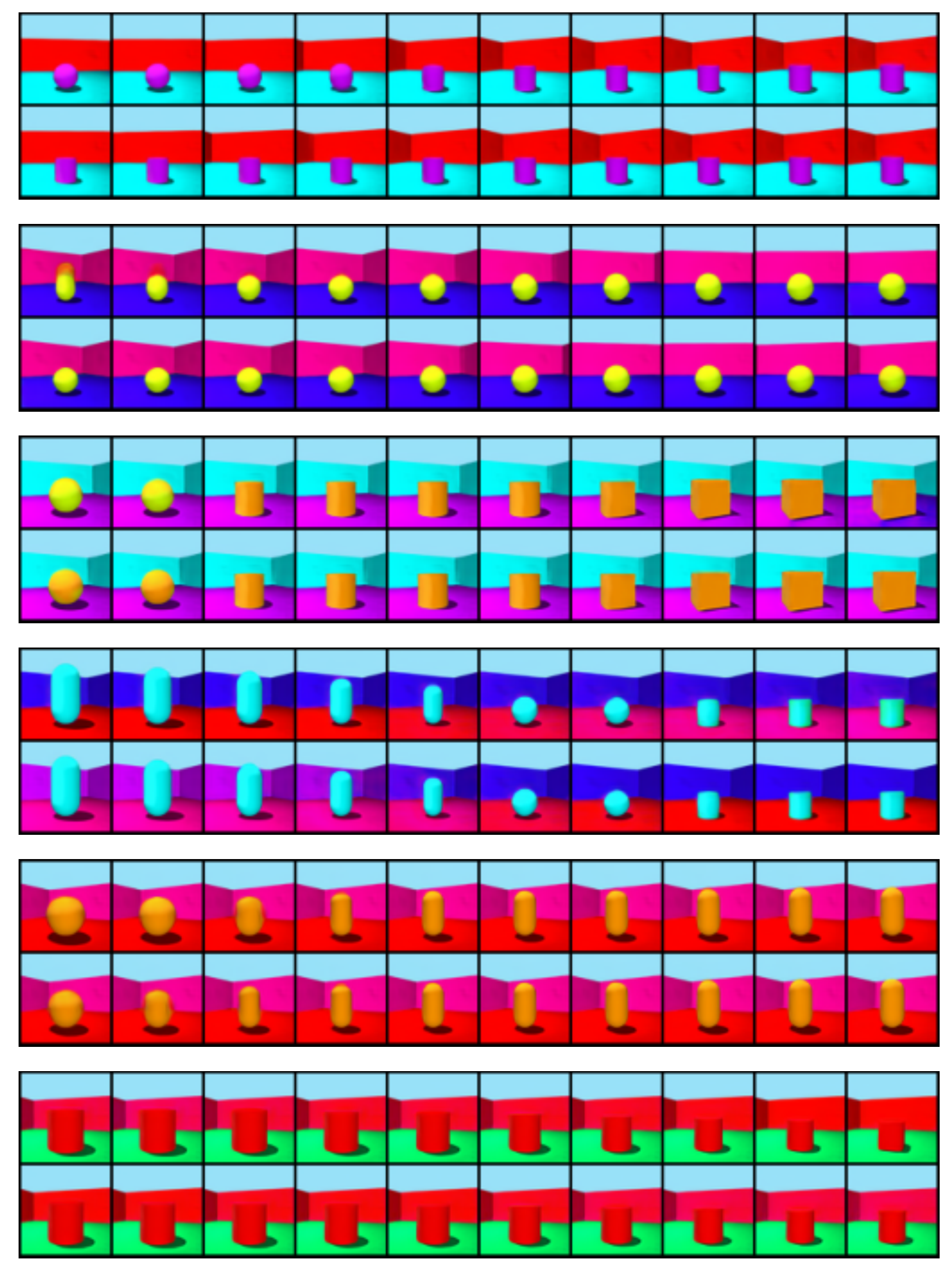} 
\caption{Comparison of latent traversals obtained by SeFa and SeFa + SRE on 3DShapes dataset. Factors shown are Orientation, Shape and Scale.}
\label{fig:appendix_shapes3d}
\end{figure*}

\begin{figure*}[t]
\centering
\includegraphics[width=0.9\textwidth]{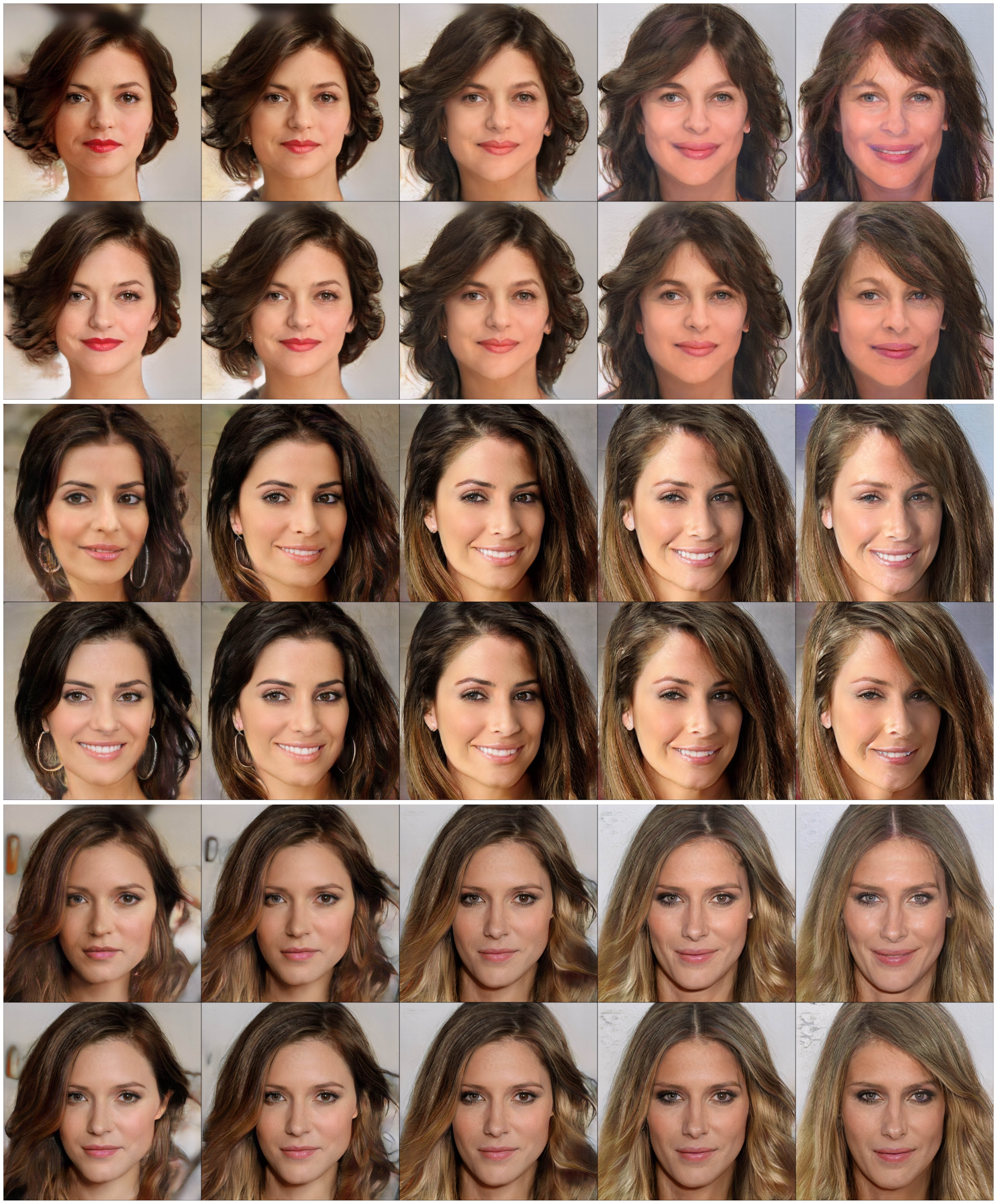} 
\caption{Comparison of latent traversals obtained by SeFa and SeFa + SRE on CelebA-HQ dataset. The attribute considered is Age.}
\label{fig:celebacf_young}
\end{figure*}

\begin{figure*}[t]
\centering
\includegraphics[width=0.9\textwidth]{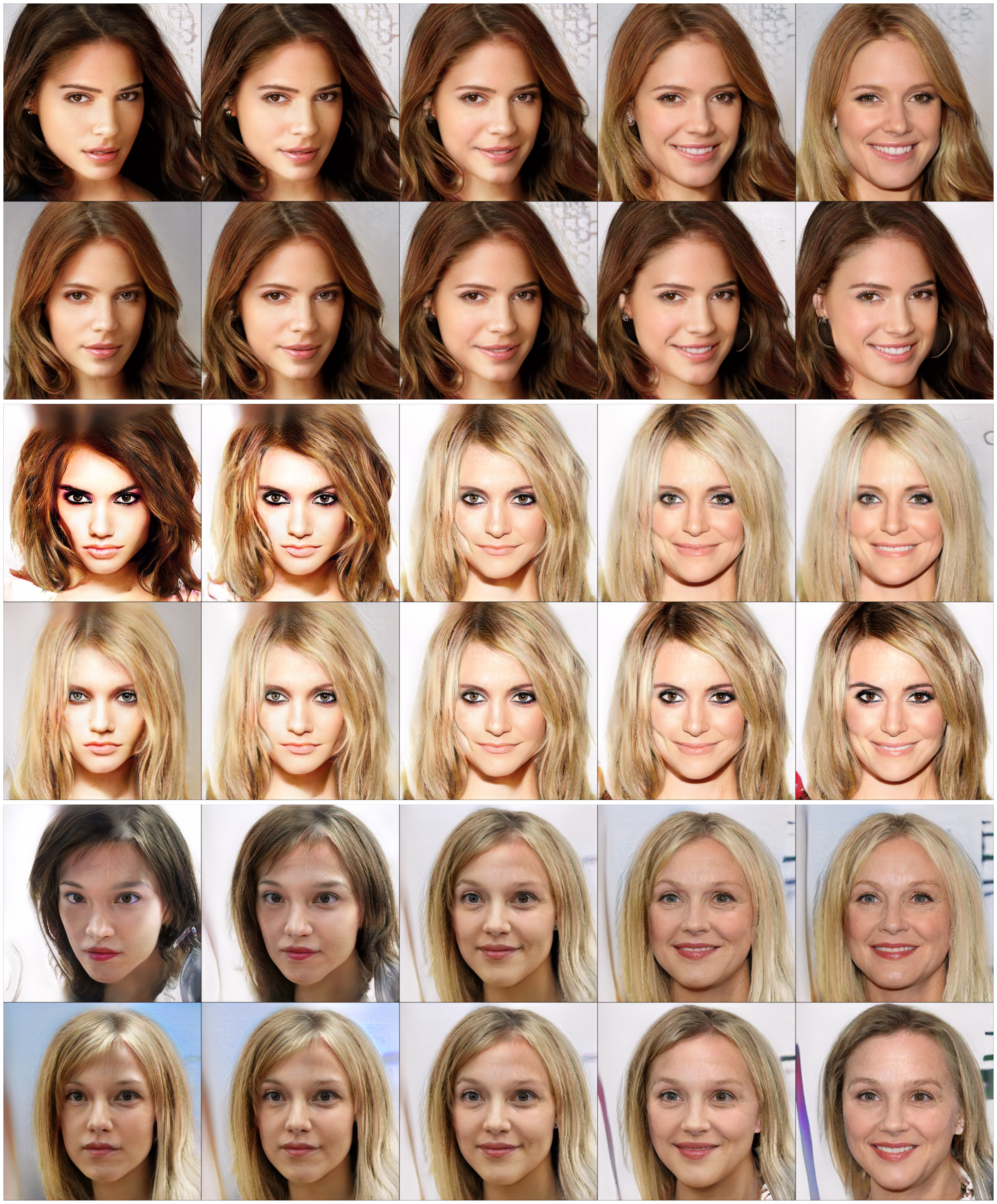} 
\caption{Comparison of latent traversals obtained by SeFa and SeFa + SRE on CelebA-HQ dataset. The attribute considered is Expression.}
\label{fig:celebacf_smiling}
\end{figure*}

\begin{figure*}[t]
\centering
\includegraphics[width=0.9\textwidth]{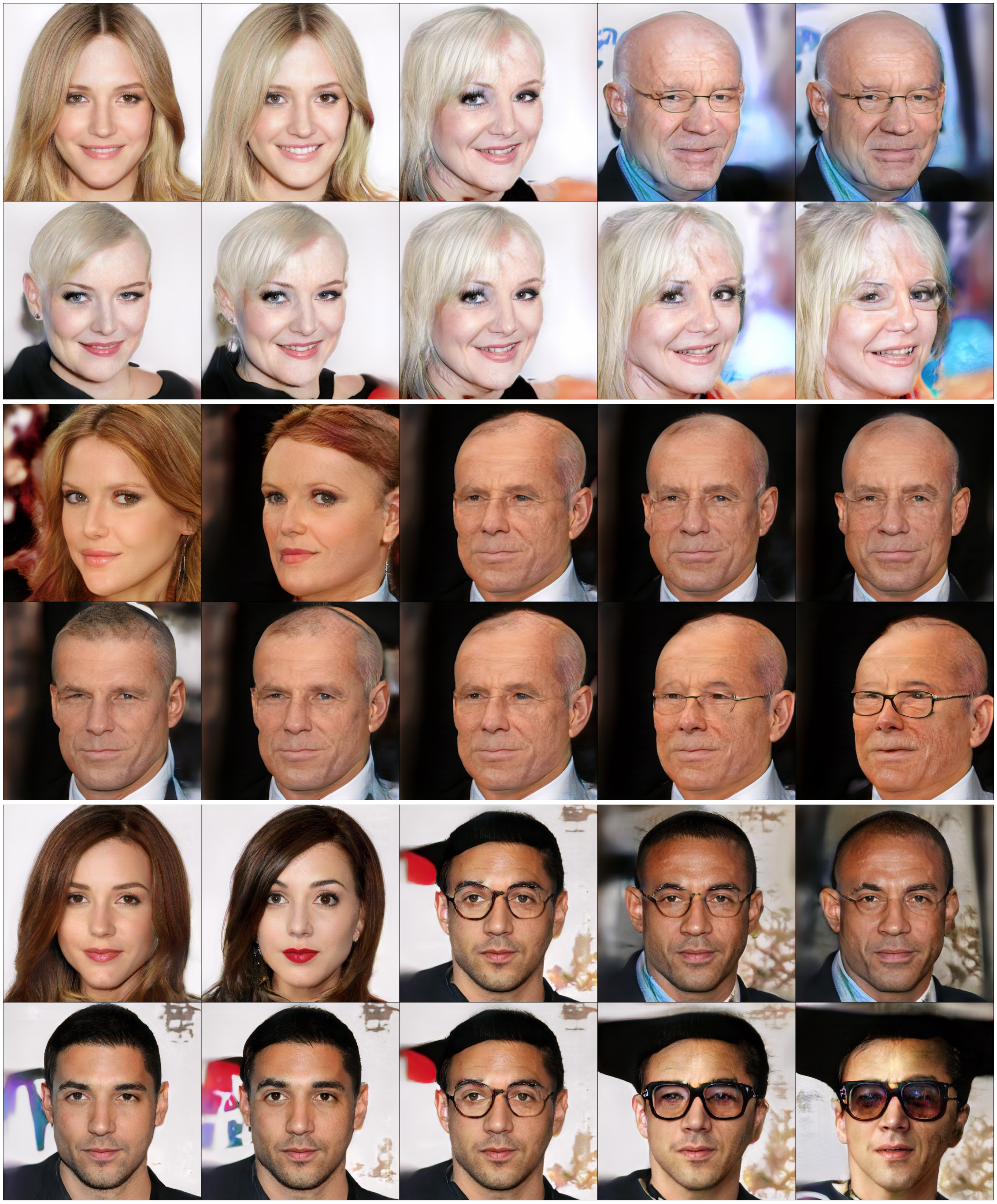} 
\caption{Comparison of latent traversals obtained by SeFa and SeFa + SRE on CelebA-HQ dataset. The attribute considered is Eye glasses.}
\label{fig:celebacf_glasses}
\end{figure*}

\begin{figure*}[t]
\centering
\includegraphics[width=0.9\textwidth]{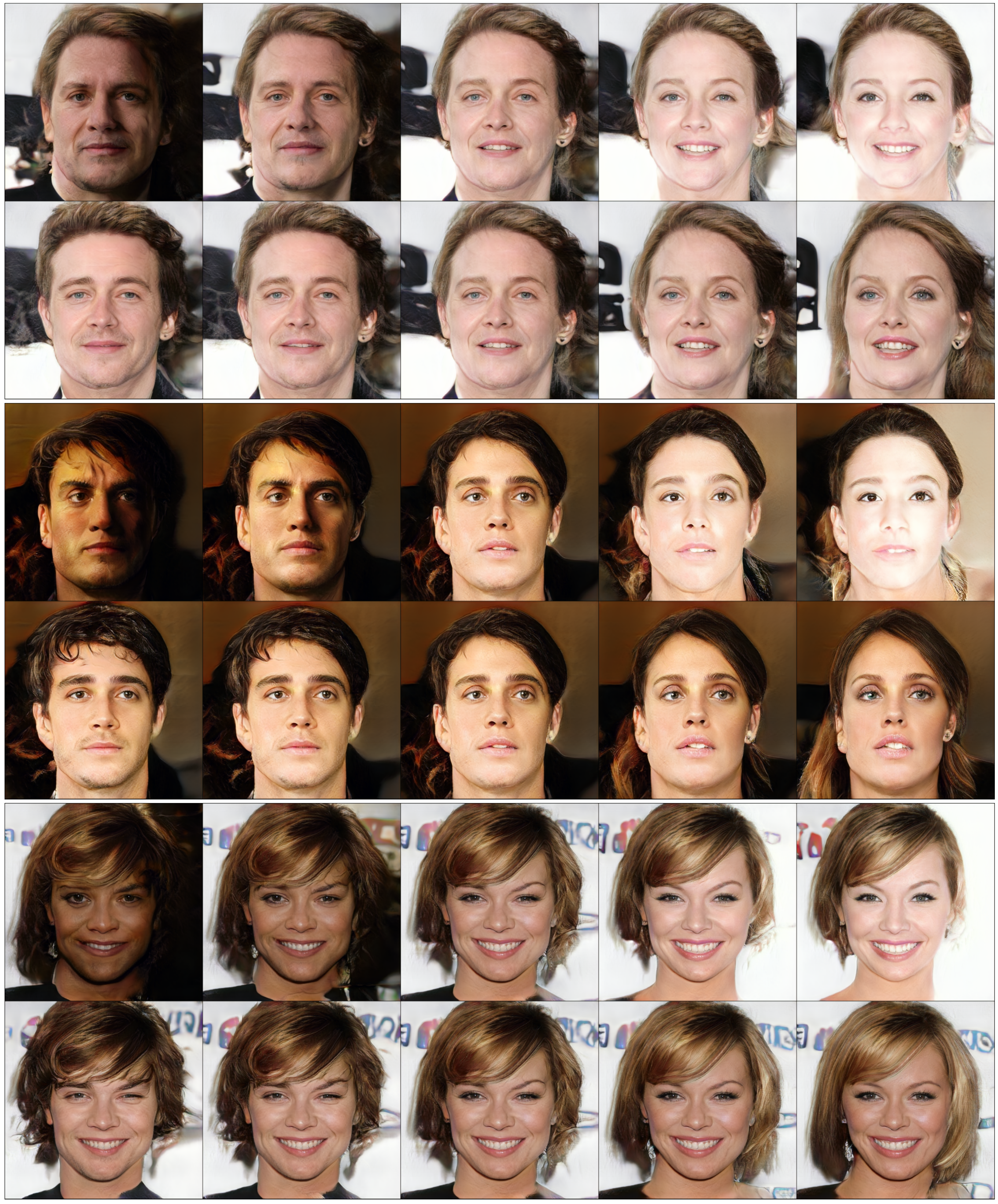} 
\caption{Comparison of latent traversals obtained by LD and LD + SRE on CelebA-HQ dataset. The attribute considered is Gender.}
\label{fig:celebald_gender}
\end{figure*}

\begin{figure*}[t]
\centering
\includegraphics[width=0.9\textwidth]{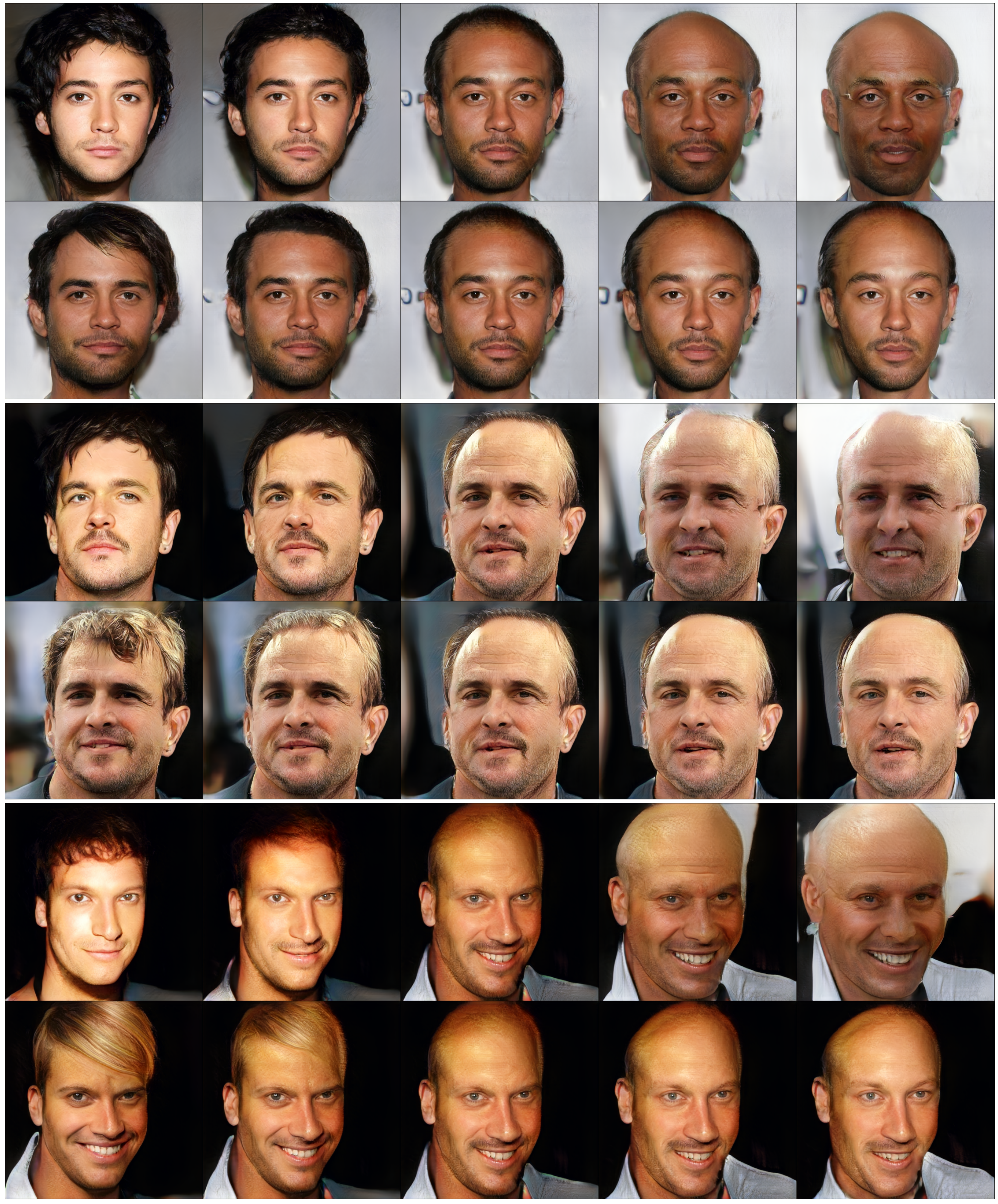} 
\caption{Comparison of latent traversals obtained by LD and LD + SRE on CelebA-HQ dataset. The attribute considered is Hair.}
\label{fig:celebald_hair}
\end{figure*}

\begin{figure*}[t]
\centering
\includegraphics[width=1.0\textwidth]{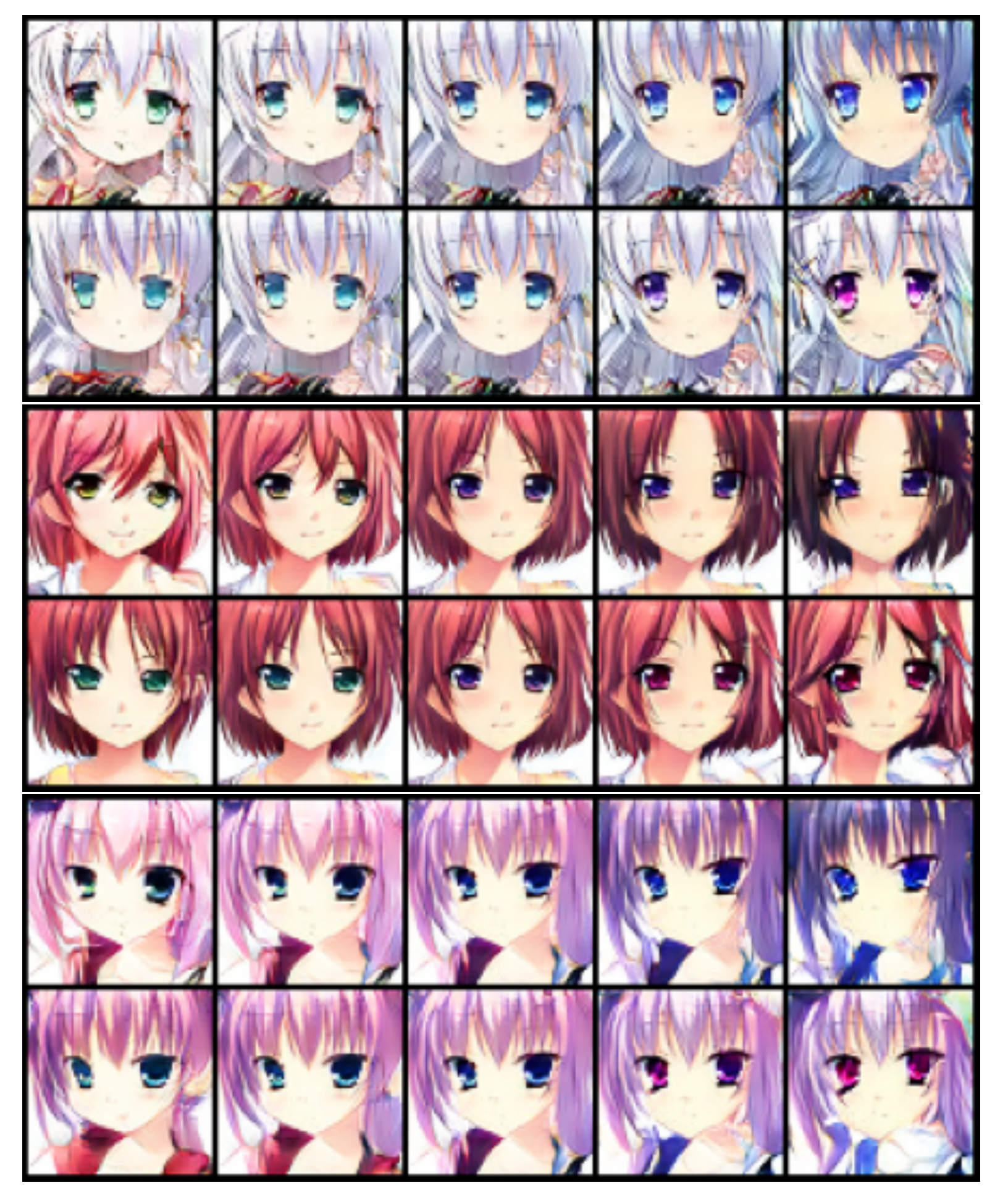} 
\caption{Comparison of latent traversals obtained by LD and LD + SRE on AnimeFaces dataset. The attribute considered is Eyecolor.}
\label{fig:anime_cf_eyecolor}
\end{figure*}

\begin{figure*}[t]
\centering
\includegraphics[width=1.0\textwidth]{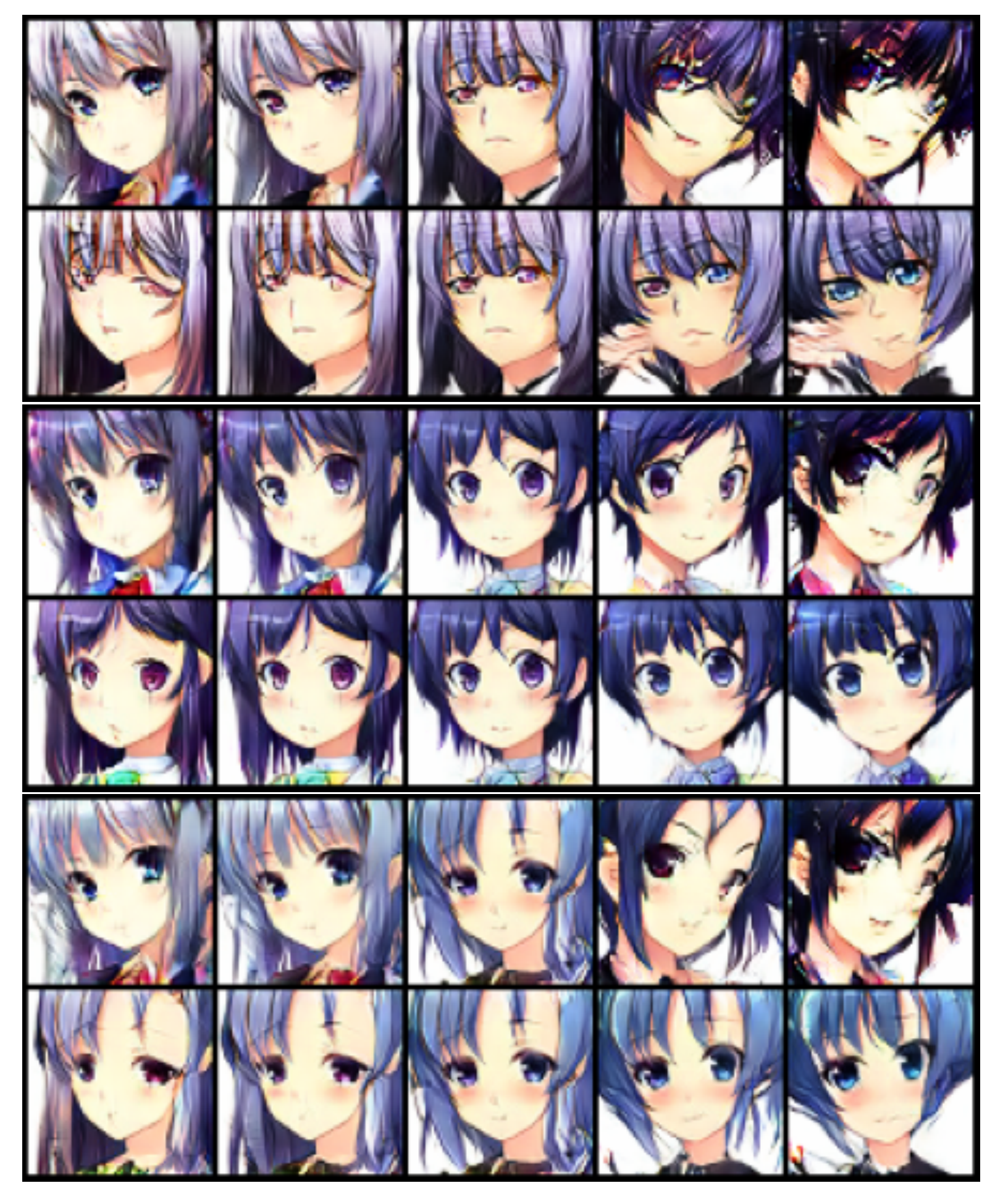} 
\caption{Comparison of latent traversals obtained by LD and LD + SRE on AnimeFaces dataset. The attribute considered is Gender.}
\label{fig:anime_cf_gender}
\end{figure*}

\begin{figure*}[t]
\centering
\includegraphics[width=1.0\textwidth]{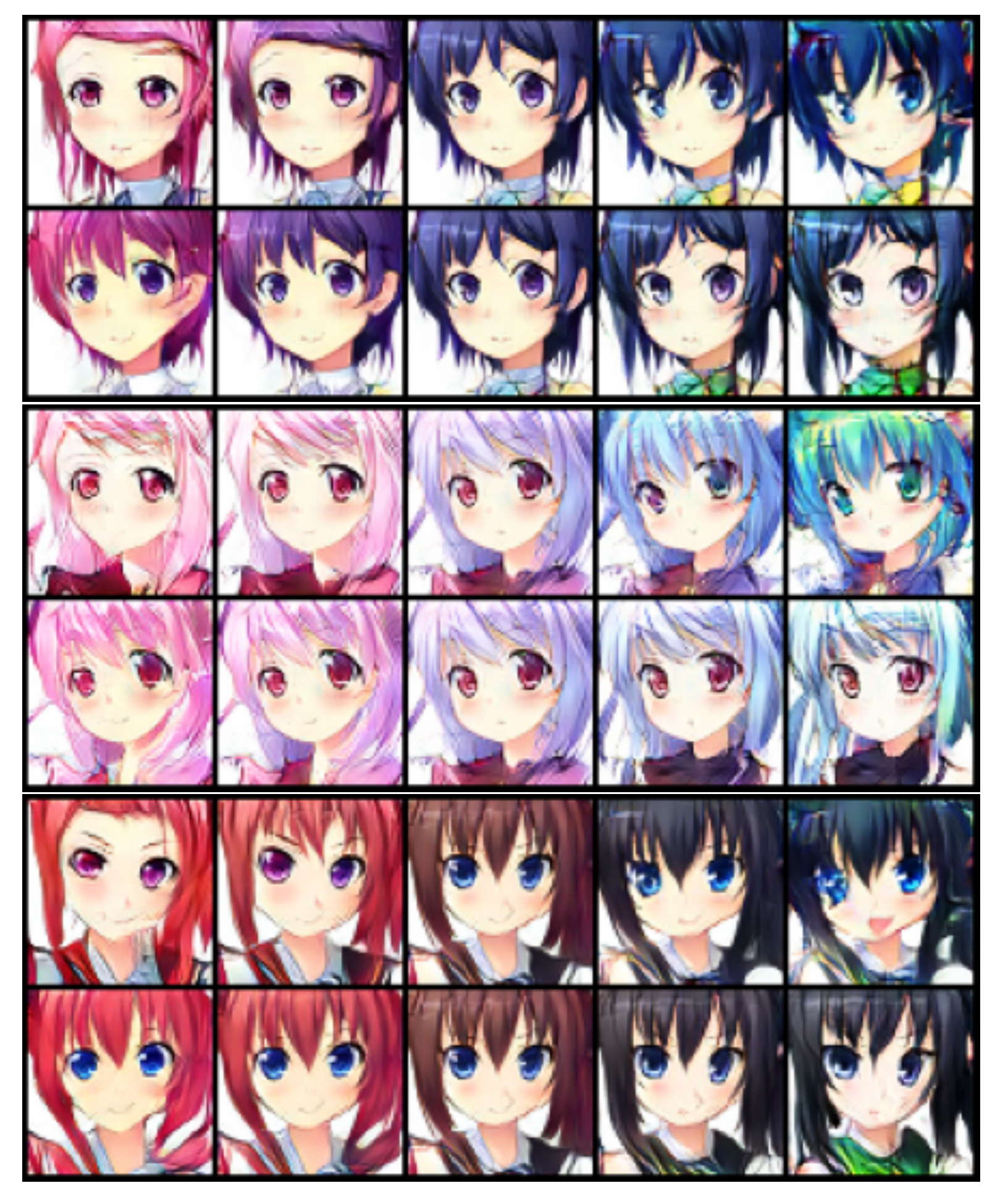} 
\caption{Comparison of latent traversals obtained by LD and LD + SRE on AnimeFaces dataset. The attribute considered is Hair color.}
\label{fig:anime_cf_haircolor}
\end{figure*}

\end{document}